\documentclass[conference]{IEEEtran}
\IEEEoverridecommandlockouts
\usepackage{cite}
\usepackage{amsmath,amssymb,amsfonts}
\usepackage{algorithmic}
\usepackage{graphicx}
\usepackage{textcomp}
\usepackage{xcolor}
\usepackage{comment}
\usepackage{todonotes}
\usepackage{subcaption}
\usepackage{caption}
\usepackage{tabularx}

\usepackage{babel}
\usepackage{url}

\usepackage{tikz}
\usetikzlibrary{arrows.meta}
\usetikzlibrary{arrows, automata,positioning,shapes}

\usepackage{balance}

\def\BibTeX{{\rm B\kern-.05em{\sc i\kern-.025em b}\kern-.08em
    T\kern-.1667em\lower.7ex\hbox{E}\kern-.125emX}}

\newcommand{\uproman}[1]{\uppercase\expandafter{\romannumeral#1}}
\newcommand{\Fig}[1]{Fig.~\ref{#1}}

\definecolor{IPA_blue}{rgb}{0.294117647,0.3294117647,0.6784313725}

\begin{document}

\title{Reinforcement Learning based Condition-oriented Maintenance Scheduling for Flow Line Systems}

\author{\IEEEauthorblockN{Raphael Lamprecht, Ferdinand Wurst}
\IEEEauthorblockA{
\textit{Center for Cyber Cognitive Intelligence (CCI),} \\
\textit{Fraunhofer IPA}\\
Stuttgart, Germany \\
raphael.lamprecht@ipa.fraunhofer.de
}

\and
\IEEEauthorblockN{ Marco F. Huber}
\IEEEauthorblockA{
\textit{Institute of Industrial Manufacturing and Management IFF,}\\
\textit{University of Stuttgart},\\
\textit{Center for Cyber Cognitive Intelligence (CCI),} \\
\textit{Fraunhofer IPA} \\
Stuttgart, Germany }
}

\maketitle

\begin{abstract}

Maintenance scheduling is a complex decision-making problem in the production domain, where a number of maintenance tasks and resources has to be assigned and scheduled to production entities in order to prevent unplanned production downtime. Intelligent maintenance strategies are required that are able to adapt to the dynamics and different conditions of production systems. The paper introduces a deep reinforcement learning approach for condition-oriented maintenance scheduling in flow line systems. Different policies are learned, analyzed and evaluated against a benchmark scheduling heuristic based on reward modelling. The evaluation of the learned policies shows that reinforcement learning based maintenance strategies meet the requirements of the presented use case and are suitable for maintenance scheduling in the shop floor.

\end{abstract}


\section{Introduction}

In cycle dependent production systems, such as flow line systems, machine breakdowns are a major risk that can lead to complete production standstills. To ensure a necessary level of availability of production entities, multiple maintenance scheduling strategies exist. On a higher level of abstraction, maintenance approaches can be grouped into strategies for corrective maintenance (CM) and preventive maintenance (PM). In CM activities are initialized after a fault or machine breakdown is detected, whereas in PM, activities are scheduled at regular time intervals with the aim of replacing components before a fault occurs \cite{Huber.2021}. In order to reduce the amount of unnecessarily executed actions, condition-based maintenance (CBM) leverages information about the current condition of the production environment. Actions are only executed if the physical condition of an asset justifies the maintenance activity~\cite{Jardine.2006}. Due to the implications of machine breakdowns, it is essential for business operators that maintenance scheduling strategies are easy to interpret, reliable, and trustworthy. Simple scheduling rules and heuristics are easy to understand and therefore are often used to schedule maintenance tasks. While these strategies are interpretable for humans, they are lacking in terms of flexibility and their capability to adapt to complex production environments. Since maintenance scheduling can be seen as a long-term optimization over a series of short-term decisions, it fits into the reinforcement learning (RL) setting, where an intelligent agent learns by interacting with an environment. Compared to simple maintenance scheduling rules, RL is highly flexible and able to leverage information of the production environment. At the same time it is not as simple and intuitive as scheduling rules and a comprehensive consideration of the behaviour of RL-based maintenance strategies is needed in order to use them in real-world production systems. 
Therefore, this work uses RL to learn condition-based maintenance scheduling strategies and evaluates their behaviour in different production settings. Our main contributions briefly are:
\begin{itemize}
    \item {Implementation of a Double Deep Q-Network for CBM scheduling.}
    \item {Formal problem description and evaluation of the influence of maintenance and production downtime costs in reward modeling for CBM.}
    \item {Comprehensive analysis and evaluation of the learned policies for maintenance scheduling.}
\end{itemize}


This paper is structured as follows: In Section \uproman{2} the fundamentals of deep RL as well as related work regarding RL for maintenance scheduling is presented. Section \uproman{3} describes the RL-based maintenance scheduling concept of this work and Section \uproman{4} illustrates the numerical results. Section \uproman{5} provides a conclusion and outlook on future research.

\section{Background}


In this section we first introduce the fundamentals of RL. Then, we review related work regarding the use of RL methods for maintenance scheduling. 

\subsection{Reinforcement Learning}

RL is a branch of machine learning where an agent learns a sequential decision problem by interacting with an environment in order to maximize a numeric reward. The decision problem can be formalized as a Markov Decision Process (MDP) with tuple $(S, A, P, R)$, where $S$ is the set of states of the environment, $A$ denotes to the set of actions and $P$ the transition probability distribution $P(s_{t+1}|s_{t},a_{t})$, which specifies for every state $s_{t} \in S$ and action $a_{t} \in A$ the probability of the next state $s_{t+1}$, where $t$ refers to a discrete time index. For every $a_{t}$ the agent receives a reward $r_{t}$, given by $R(s_{t}, a_{t}, s_{t+1})$. Through the sequential interaction of the agent in the MDP, a series of subsequent tuples $(s_{t}, a_{t}, r_{t}, s_{t+1})$ form a trajectory $\tau = (s_0, a_0, r_0, s_{1}, ... , s_T, a_T, r_{T}, s_{T+1})$.
The return of a trajectory $R = \sum_{k=0}^{T} \gamma^k \cdot r(s_k, a_k)$ is the discounted sum of rewards received by the agent during the interaction, where $\gamma \in [0,1]$ is a discount factor that scales the importance between intermediate and future rewards. The goal of RL is to learn a policy $\pi : S \mapsto A$, which maximizes the expected return over all trajectories \cite{Sutton.2018}. If the transition probability function $P$ is known, the optimal policy $\pi^*$ can be obtained by dynamic programming \cite{Bertsekas.2016}. Since the prerequisite of a known $P$ is often not satisfied, RL allows learning a policy without knowing $P$. Often used RL methods can be categorized into two groups: value-based and policy-based methods. Value-based methods aim to learn an approximation of the Q-function, which maps every state-action pair $(s_{t},a_{t})$ to the expected future reward. The policy is then derived by acting greedily with respect to the estimated Q-values, i.e., 
\[
    \label{policy_Q}
    \pi(s_{t})_{Q} = \operatorname*{arg} \operatorname*{max}_a  Q(s_{t},a_{t})~.
\]
On the other hand, policy-based methods try to directly learn the policy without formulating a Q-function. Therefore, a policy $\pi(s_{t})_\theta$ is described by a parametric function and optimized using optimization methods such as the gradient descent algorithm. Modelling of $Q$ and $\pi$ is an essential task in RL. If the state-action space $|S|\times |A|$ is sufficiently small, tabular methods can be used to store a value for every state-action pair $(s_{t},a_{t})$. As the required size of the table grows rapidly with an increase of $S$ and $A$, their use is restricted to problems with limited complexity. In more complex settings it is common to use parametric models with parameters $\hat \theta$ with $\dim(\hat \theta) \ll |S| \times |A|$ for representing $Q$ and $\pi$. Over the last years a lot of research has been put into the usage of deep neural networks (DNNs) as function approximators for $Q$ and $\pi$ \cite{Arulkumaran.2017, Lazaridis.2020}. DNNs enable RL to scale to problems with high-dimensional state and action spaces and produced remarkable success stories. Work conducted by \cite{Mnih.2013} demonstrated that a variant of Q-Learning, the Deep Q-network (DQN), could learn to play a collection of Atari games by using the screen images as input. In continuous environments policy-based algorithms with DNNs, such as Proximal Policy Optimization (PPO)\cite{Schulman.2017}, have shown remarkable results in domains like robotics and navigation \cite{Arulkumaran.2017}.


\subsection{RL in Maintenance Scheduling}


The interest of the manufacturing research community in the application of RL to solve complex decision problems in the area of production has increased rapidly in recent years. In the area of maintenance planning, \cite{Wang.2014, Wang.2016, Zheng.2017, Ling.2018, Knowles.2011} are using tabular Q-learning to learn maintenance policies for single production machines or two machine systems with one intermediate buffer. Whereas \cite{Wang.2014} and \cite{Knowles.2011} are only considering two actions (maintenance, no maintenance), \cite{Wang.2016} and \cite{Ling.2018} differentiate between corrective and preventive maintenance. \cite{Huang.2019} is using a tabular Q-learning algorithm to learn a preventive maintenance strategy for a small flow line system consisting of three machines with two intermediate buffers.

Since tabular methods suffer from the curse of dimensionality, their use is impractical for production settings aiming to address real-world applications. Tabular methods cannot be applied to more complex productions systems. This work therefore addresses the use of function approximators to learn condition-oriented maintenance policies for flow line systems, which are scalable to problems in high-dimensional state and action spaces. Since another major challenge in real-world RL applications is the explainability of policies \cite{DulacArnold.2019}, this work focuses on the analysis of the learned policies and their applicability in different production settings. This work makes a contribution towards developing effective, robust and trustworthy policies in complex production settings with high-dimensional state and action spaces.

RL with function approximation in the context of a job shop is also considered by \cite{Kuhnle.2019}. The authors are using PPO for maintenance planning of parallel, single machines. Other work by \cite{Huang.2020} is considering inter machine dependencies in a flow line system using value-based Double Deep Q-Network (DDQN) algorithm. The production system is modeled as a dynamic system in a state-space representation, described by \cite{Zou.2018}.
Instead of a state-space representation, our work uses a discrete event simulation for modeling the production environment, which is more flexible in modeling complex systems and taking into account stochastic elements.



\section{RL-Agent Formulation}


In this section the methodology for using RL for condition-oriented maintenance scheduling in flow line systems is described.

\subsection{Production System}

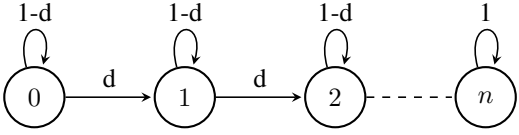
\begin{figure}[tb]
    \centering
    \begin{tikzpicture}[
        > = stealth, 
        shorten > = 1pt, 
        auto,
        node distance = 2 cm, 
        semithick 
    ]
    \tikzstyle{every state}=[
        draw = black,
        thick,
        fill = white,
        minimum size = 8mm
    ]
    \node[state] (0) {$0$};
    \node[state] (1) [right of = 0] {$1$};
    \node[state] (2) [right of = 1] {$2$};
    \node[state] (n) [right of = 2]{$n$};
    
    \path[loop above] (0) edge node {1-d} (0);
    \path[->] (0) edge node {d} (1);
    \path[loop above] (1) edge node {1-d} (1);
    \path[->] (1) edge node {d} (2);
    \path[loop above] (2) edge node {1-d} (2);
    \path[dashed] (2) edge node {} (n);
    \path[loop above] (n) edge node {1} (n);

\end{tikzpicture}
    \caption{Markov process of the machine degradation.}
    \label{degradation_process}
\end{figure}

This work considers a flow line system consisting of $i$ different machines $M= \{m_{1},m_{2}, ..., m_{i}\}$ with intermediate buffers. Each machine is described by a unique process time $p$, degradation rate $d$ and a buffer size $b$ of the upstream buffer. The production system is modelled as a discrete event simulation with simulation time $t_{\mathrm{sim}}$. When a machine is operational, it takes a part from the upstream buffer and starts processing it. While processing, the machines can degrade. The degradation process is formulated as a discrete Markov process, shown in \Fig{degradation_process}. Each machine has $n+1$ different states $cs \in \{0,1, ...,n\}$, representing the condition of the machine. At the beginning of the simulation, every machine starts with $cs=0$, indicating the best possible condition. The transition probability with which a machine changes from its current condition state $h$ to $h+1$ is given by the degradation rate $d$. The degradation process is triggered at each simulation time step in which the machine is in operation. If a machine reaches state $n$ it breaks down. The Markov process is used within the simulation environment to obtain subsequent states and is not explicitly used as part of a dynamics model of the production system to determine the optimal strategy via dynamic programming. Since dynamic programming requires a tabular representation, it is not applicable for larger $S$ and~$A$. Investigating model-based RL to use known dynamics models for planning purposes is part of future research. Before reaching the breakdown state, machines can receive CBM. If the machine is already in state $n$, only CM can be performed to get the machine working again. After processing a part by a machine, the part is put into the downstream buffer, where it remains until the downstream machine can process it. For modelling the production system the discrete event simulation package SimPy is used.

\subsection{Agent}

The goal of the RL agent is to schedule maintenance activities for the previous described production system in order to maximize a production specific objective function. Since the availability of maintenance resources is constrained in practice, we do not allow parallel execution of maintenance activities. The term maintenance resource is used to indicate the availability of both, a skilled maintenance employee and all necessary technical equipment such as spare parts to successfully carry out the job.
To control the interaction of the agent with the production environment, so-called decision points are defined. If a decision point occurs, the simulation stops and does not start again until the agent has performed an action. The simulation then continues to run until the next decision point is reached. A decision point occurs if two conditions are fulfilled: (i) a maintenance resource is available and (ii) the condition state of a machine is above a critical state $n_c$. The critical state of a machine is introduced to benchmark the RL policies against a FIFO-policy (First in First Out), where the machines are maintained in the sequence in which they request maintenance. The value for $n_c$ is defined empirically. For training the RL policies, the threshold state is set to $n_c=0$ to allow the greatest possible scope for decision-making. Therefore, only condition (i) is relevant in the RL setting. At a decision point, the agent can either perform a maintenance action on one of the machines $m \in M$ or choose the idle action, where nothing is done. For $i$ machines the size of the action space is thereby given by $i+1$. The state representation of the MDP thus consists of the value of the current condition for every machine $m \in M$ and the buffer levels for all machines. With machine specific buffer sizes $b_{j}$ the size of $S$ is given by $\prod_{j=1}^{i} ((n+1) \cdot b_{j})$.

For the implementation of the agent, a DDQN \cite{vanHasselt.2016} is used. A DDQN is a variant of a DQN with a target network and experience replay \cite{Mnih.2015} but slightly differs in the way actions are evaluated and selected. Compared to a DQN, the values of the target network $Q(s_{t}, a_{t} ;\theta^{\prime})$ are only used for the evaluation of the current greedy action. For the policy, the values obtained by the DQN network $Q(s_{t},a_{t}; \theta)$ are instead used to select the action. This procedure reduces the frequent problem of overestimating the Q-values due to estimation errors \cite{vanHasselt.2016}. The target Q-value $Q^*$ for the update procedure $Q(s_{t},a_{t}, \theta)~\xleftarrow{}~Q(s_{t},a_{t},\theta) + \alpha \cdot (Q^{*}-Q(s_{t},a_{t},\theta))$ is given by
\[
    \label{DDQN_update}
    Q^{*} = r_{t+1} + \gamma  \cdot Q(s_{t+1}, \operatorname*{arg} \operatorname*{max}_a Q(s_{t+1},a_{t}; \theta); \theta^{\prime})
\]for the DDQN. DQNs and its variations such as the DDQN have shown great achievements in various environments with discrete state and action spaces, such as Atari games \cite{Mnih.2015} and various applications in the production domain \cite{Waschneck.2018}. Since the formulated problem in this work is modelled as a discrete environment with discrete state and action space, a DDQN is used for the agent implementation.

\subsection{Reward Design}

The goal of maintenance scheduling is to ensure a high level of machine availability with using as little maintenance related costs as possible. Since ensuring a high level of availability of the production system comes with the cost of regularly performing maintenance activities, it often conflicts with the short-term operational goals of high utilization in the shop floor. These considerations must be taken into account when designing the reward for the RL agent. Many different reward functions are possible based on the operational focus. For the purpose of this paper, two different reward functions are defined and compared. Reward
\[
    \label{eq:reward_1}
    R_1 = pp_{\mathrm{t_{k+1}}} - pp_{\mathrm{t_{k}}}
\]
only focuses on the output quantity of the production system. After every action the intermediate reward is given by the difference between the produced parts $pp$ at the simulation time $t_{k}$ of the current decision point $k$ and the following one $t_{k+1}$. In contrast to $R_1$, reward function $R_2$ is more sophisticated and also considers the maintenance costs besides the output of the production system.
\[
    R_2 = \left.
    \begin{cases}
        0 & \text{for scenario A}\\
        -\left(\beta \cdot c_{\mathrm{CBM}}\right) & \text{for scenario B} \\
        -\left(c_{\mathrm{CBM}} + c_{\mathrm{CM}} + \dfrac{c_{\mathrm{pl}}}{t_{k+1} - t_{k}} \right) & \text{for scenario C}
    \end{cases}
    \right.
\]
is based on three different scenarios (scenario A-C) and comprises three different cost terms: the CM costs $c_{\mathrm{CM}}$, the CBM costs $c_{\mathrm{CBM}}$ and a cost factor for the production loss $c_{\mathrm{pl}}$, which is scaled by the simulation duration $t_{k+1} - t_{k}$ between the current decision point $k$ and the next decision point $k+1$. The scaling is done to prevent disproportionate penalization for maintenance actions due to the production downtime costs incurred in each simulation step. Without this scaling, there is a risk that the agent will wait only to avoid being penalized. $\beta > 0$ is a penalty factor. Scenario A and B occur if the agent chooses the idle action and does not perform maintenance. Since the idle action is not favorable if machines are broken, the agent gets a negative reward of $-(\beta \cdot c_{\mathrm{CBM}})$ if at least one machine is in the breakdown state $n$ (scenario~B) but does not get punished if none of the machines are broken down at the time the next decision point is reached (scenario~A). Scenario~C covers the case when the agent schedules either CBM or CM actions. The reward is calculated according to \cite{Huang.2019} as a cost function, considering cost factors $c_{\mathrm{CM}}$, $c_{\mathrm{CBM}}$, and  $c_{\mathrm{pl}}$. The values for the cost factors and penalty factor are chosen iteratively by comprehensive experiments and set to $c_{\mathrm{CBM}}= 0.5$, $c_{\mathrm{CM}}=1.5$, $c_{\mathrm{pl}}=0.1$ and $\beta=10$ in the following experiment.

\section{Experiments}

\begin{figure}[tb]
	\centering
    \includegraphics{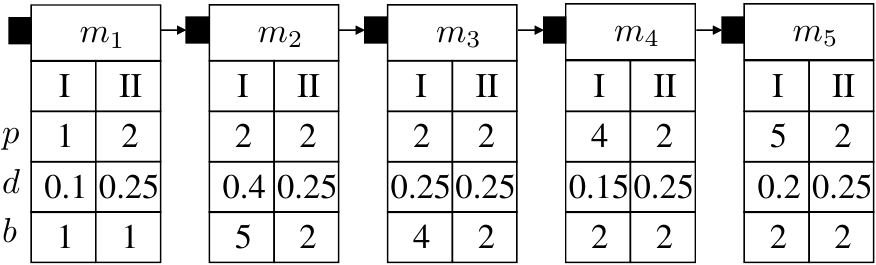}
	\caption{Production system with asynchronous configuration~\uproman{1} and synchronous configuration~\uproman{2}.}
	\label{Fig31}
\end{figure}

In order to show the applicability of the methodology to a wider range of flow line systems, the production system is implemented in two different configurations with $i=5$ machines, shown in \Fig{Fig31}. In configuration~\uproman{1} the machines are characterized by high variance in their process times $p$. Also the degradation rates $d$ and buffer levels $b$ differ significantly between the machines. In contrast to configuration~\uproman{1}, a classical, synchronized flow line system with low variation in $p$, $d$ and $b$ is considered in configuration~\uproman{2}. The values for $p$, $d$ and $b$ are chosen in such a way that they fit to the duration of the maintenance activities and degradation occurs at moderate simulation times. The duration of the maintenance actions are adapted from \cite{Huang.2019} and set to $t_{CM}=20$ simulation time steps for CM and $t_{CBM}=5$ for CBM. The duration differs because preventive maintenance usually can be performed faster while CM takes longer since it is more likely that the reason for breakdown is unknown or necessary equipment may not be in stock. The duration of the idle action is one simulation time step. The breakdown state is set to $n=10$. The threshold for the critical state $cn_c$ for the FIFO-policy is determined empirically by performing trials with all possible values of the threshold $cn_c \in \{0,1, ...,n\}$. The threshold that leads to the highest production rate or lowest maintenance cost is then used for the FIFO-policy. For configuration \uproman{1}, the best performing threshold is $cn_c=6$ and $cn_c=5$ for configuration \uproman{2}.


\subsection{Results}

\begin{figure}[t]
    \centering
    \includegraphics[width=\linewidth]{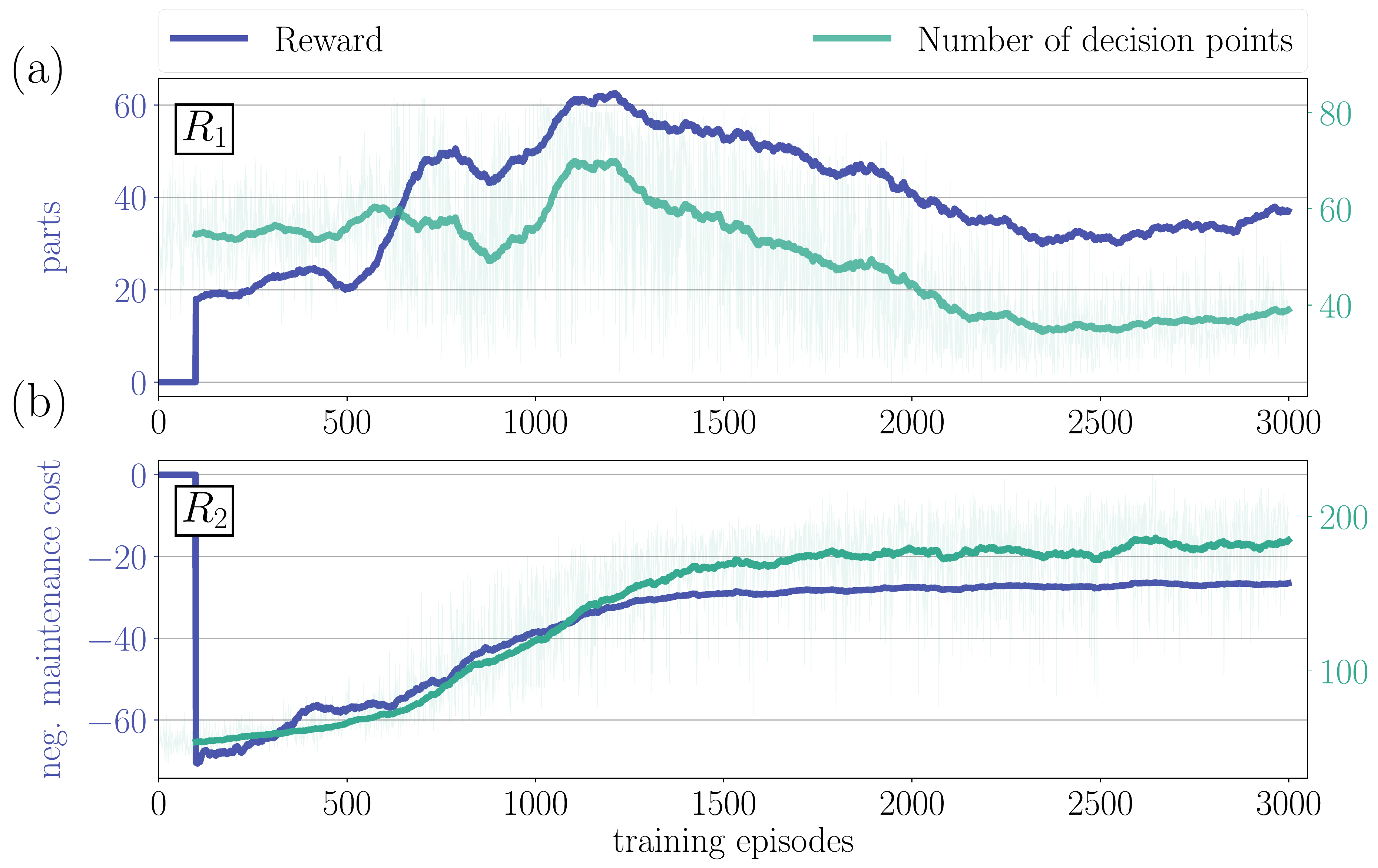}
     \caption{Reward and number of decision points over the training with (a) $R_{1}$ and (b) $R_{2}$ for configuration~\uproman{1}.}
     \label{training_DDQN}
\end{figure}

In order to train the DDQN algorithm, 3,000 episodes are performed. In every episode the simulation time is set to $t_{\mathrm{sim}}=400$. The training results for configuration~\uproman{1} are shown in \Fig{training_DDQN}, where the total number of produced parts and the generated maintenance costs per simulation run over the training episodes are smoothed over 100 episodes.

For the DQN network $Q(\theta)$ and the target network $Q(\theta^{\prime})$ fully connected DNNs with two hidden layers and relu activation are used. The hyperparameters (HP) of the DDQNs are selected using Bayesian optimization with a Gaussian process as surrogate for the objective function. As objective, the maximum total reward smoothed over 100 episodes is optimized. The optimization is performed for both reward functions and therefore two DDQNs with different HP are used for comparison. For $R_{1}$ the number of neurons of the hidden layers is set to 17 and 11 and for $R_{2}$ to 14 and 18. The size of the replay memory is set to 100,000 for both DDQNs with a batch size of 151 in the $R_{1}$ setting and 137 in the $R_{2}$ setting. During training, the parameters $\theta^{\prime}$ are updated every 200 episodes when training $R_{1}$ and every 98 episodes when training $R_{2}$. $\gamma$ is set to 0.870 for $R_{1}$ and to 0.993 for $R_{2}$. As optimization algorithm, adam is used to minimize the mean square error loss with the configuration described in~\cite{Kingma.2015}. The initial learning rate is set to $5.4\times10^{-4}$ in $R_{1}$ and $3.6\times10^{-4}$ in $R_{2}$. To ensure exploration during the training phase, the $\epsilon$-greedy policy is used. The value of $\epsilon$ describes the probability with which the agent executes a random action instead of acting greedily with respect to $Q(s_{t},a_{t})$. At the beginning of the training the agent starts with $\epsilon = 1$ and then decays $\epsilon$ by a factor of $4.8\times10^{-5}$ every episode in $R_{1}$ and $2.9\times10^{-5}$ in $R_{2}$ until a minimum of $\epsilon = 0.1$ is reached. During training with $R_{1}$, the performance drops after around 1,300 episodes, see \Fig{training_DDQN}(a). This phenomenon is known as catastrophic forgetting. To avoid negative implications by catastrophic forgetting, the model parameters are constantly saved and only updated if the performance increases. In the training of $R_{2}$, catastrophic forgetting does not occur. The reward constantly increases and finally converges after around 1,500 episodes. It is noticeable that the number of decision points or performed actions by the agent correlates with the performance of the agent. This is due to the increased selection of the idle action when the agent learns to wait for the optimal time for performing maintenance actions. The training of the polices for configuration~\uproman{2} is carried out in the same manner using the same HP as for the DDQN trained for configuration~\uproman{1}.

\begin{figure}[tb]
    \centering
    \includegraphics[width=\linewidth]{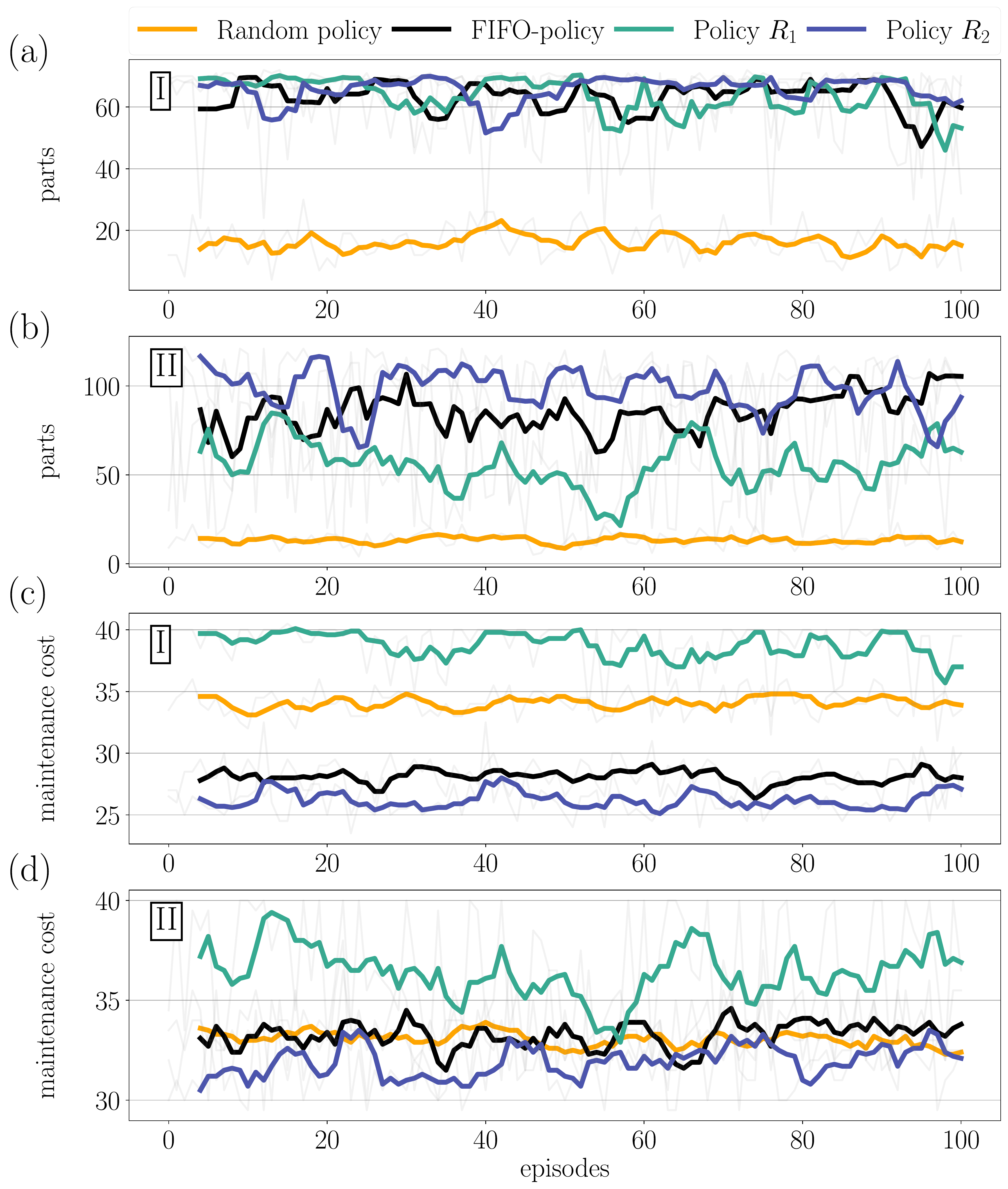}
     \caption{(a), (b) produced parts and (c), (d) maintenance costs of the considered maintenance strategies over 100 episodes for configuration \uproman{1} and configuration~\uproman{2}.}
     \label{pp_mc}
\end{figure}

To compare the learned policies, they are applied for maintenance scheduling in both configurations over 100 different simulation episodes. As a benchmark, the FIFO-policy and a random policy are additionally applied. \Fig{pp_mc} shows for every episode in (a), (b) the total number of produced parts and in (c), (d) total maintenance costs for configuration~\uproman{1} and configuration~\uproman{2}. The average performance over all 100 episodes and the resulting production rates are shown in Table~\ref{table:1}. Assuming ideal machine conditions without degradation, the maximum output quantity of the system configurations are calculated by $\rho_\mathrm{max} = t_\mathrm{sim}/p_\mathrm{max}$. In configuration~\uproman{1} the longest machine process time is $p_\mathrm{max} = 5$ and therefore $\rho_\mathrm{max} = 80$, while in configuration~\uproman{2} $p_\mathrm{max} = 2$ and $\rho_\mathrm{max} = 200$. For all three metrics the policy $R_{2}$ achieves the best results in both configurations, see Table~\ref{table:1}. Policy $R_{1}$ shows two main disadvantages. First, while it acts sufficiently well in configuration~\uproman{1}, it acts poorly in the synchronized flow line configuration \uproman{2}. Second, the associated maintenance costs are the highest of all four policies. This behavior is due to the simple implementation of the reward function, which does not consider costs.

\begin{table}[tb]
    \caption{Comparison of the performance for different strategies in configuration~\uproman{1} and \uproman{2}.}
    \begin{tabularx}{\linewidth}{c c c c c c c c}
        \hline
        \multicolumn{2}{l}{} & \multicolumn{2}{l}{Production rate} &\multicolumn{2}{l}{\#parts}& \multicolumn{2}{l}{Maintenance costs}\\
        \hline
        \multicolumn{2}{l}{Configuration} & \uproman{1} & \uproman{2} & \uproman{1} & \uproman{2} & \uproman{1} & \uproman{2} \\
        \hline
        \multicolumn{2}{l}{Policy $R_1$} & 79.8\% & 28.0\% & 63.8 & 56.0 & 38.7 & 36.4 \\
        \multicolumn{2}{l}{Policy $R_2$} & \textbf{81.2\%} & \textbf{49.1\%} & \textbf{65.4} & \textbf{98.1} & \textbf{26.2} & \textbf{31.9} \\
        
        \multicolumn{2}{l}{FIFO-policy} & 78.9\% & 42.8\% & 63.1 & 85.5 & 28.1 & 33.3 \\
        
        \multicolumn{2}{l}{Random policy} & 20.0\% & 6.7\% & 16.0 & 13.3 & 34.6 & 33.1 \\
        \hline
    \end{tabularx}
    \label{table:1}
\end{table}

\subsection{Policy Evaluation}

A major challenge for the use of RL in real-world applications is the desire of system operators for explainable policies and actions \cite{DulacArnold.2019}. Maintenance scheduling is nowadays performed by human operators who are using experiential knowledge and easy to understand heuristics for their decision-making. To use RL systems for maintenance scheduling, it is essential that human operators are able to understand the behavior of an autonomously learned policy. This is especially relevant if the system might find an alternative or unexpected approach to solve the scheduling task. To evaluate the behavior of the learned policies, they are evaluated in two dimensions: \emph{number of maintenance actions} that are performed and their \emph{timing}.

\subsubsection{Number of Maintenance Actions}

\begin{figure}[tb]
    \centering
    \includegraphics[width=\linewidth]{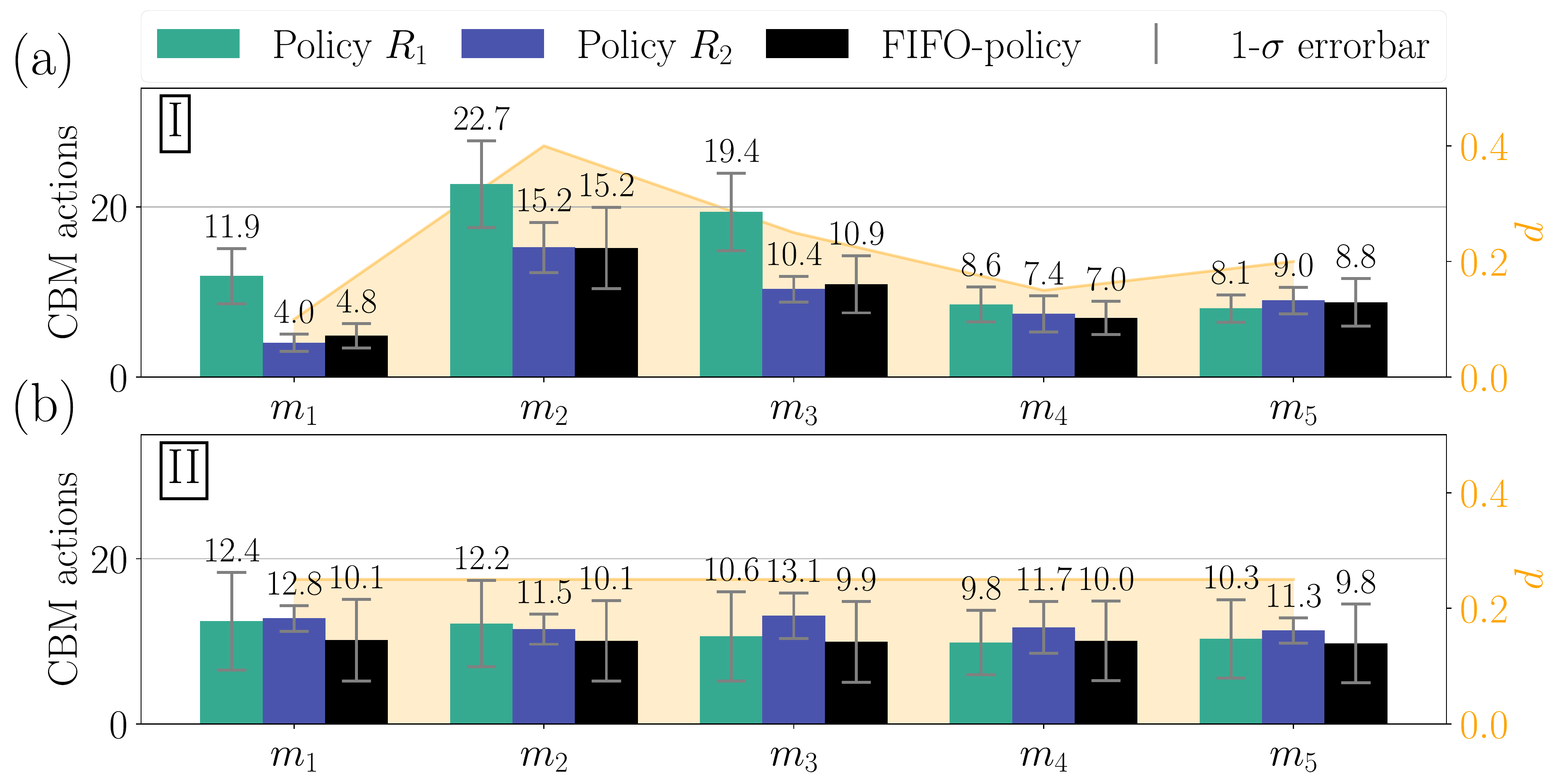}
     \caption{Machine specific CBM actions per episode in (a) configuration~\uproman{1} and (b) configuration~\uproman{2}.}
     \label{CBMtotal}
\end{figure}

\begin{table}[b]
    \centering
    \caption{Average number of CBM, CM and idle actions per episode.}
    \begin{tabularx}{0.82\linewidth}{c c c c c c c c}
        \hline
        \multicolumn{2}{l}{} & \multicolumn{2}{l}{\#CBM} &\multicolumn{2}{l}{\#CM} & \multicolumn{2}{l}{\#idle}\\
        \hline
        \multicolumn{2}{l}{Configuration} & \uproman{1} & \uproman{2} & \uproman{1} & \uproman{2} & \uproman{1} & \uproman{2}\\
        \hline
        \multicolumn{2}{l}{Policy $R_1$}    & 70.6  & 55.3  & 2.4   & 6.4 & 0 & 0\\
        \multicolumn{2}{l}{Policy $R_2$}     & 46.1  & 60.3  & \textbf{2.2}   & \textbf{1.0} & 127.7 & 79.1\\
        \multicolumn{2}{l}{FIFO-policy}     & 46.7  & 49.9  & 3.0   & 5.5 & 102.4 & 48.8\\

        \hline
    \end{tabularx}
    \label{table:2}
\end{table}

Technical components and production machines vary in their useful life and vulnerability for breakdowns. For distributing limited maintenance resources this has to be taken into account by maintenance scheduling policies. In configuration \uproman{1} the machines have different degradation rates, whereas in configuration \uproman{2} all machines degrade with a rate of $d=0.25$. \Fig{CBMtotal} shows that for the policies $R_{1}$, $R_{2}$ and FIFO the average number of performed CBM actions per simulation run correlates with the degradation rate of the individual machines. The number of CBM actions with policy $R_{1}$ is significantly higher as for $R_{2}$ and the FIFO-policy. Since the associated costs are not implicitly considered in $R_{1}$, this behavior is expected. On the other hand policy $R_{2}$ encounters maintenance associated costs directly and does not show this behavior. In conclusion, policy $R_{2}$ achieves the highest production rate with generating the lowest costs at the same time, see Table~\ref{table:1}. Moreover, $R_{2}$  has the lowest average number of unwanted CM actions, see Table~\ref{table:2}.

\subsubsection{Timing of Maintenance Actions}

\begin{figure}[tb]
    \centering
    \includegraphics[width=\linewidth]{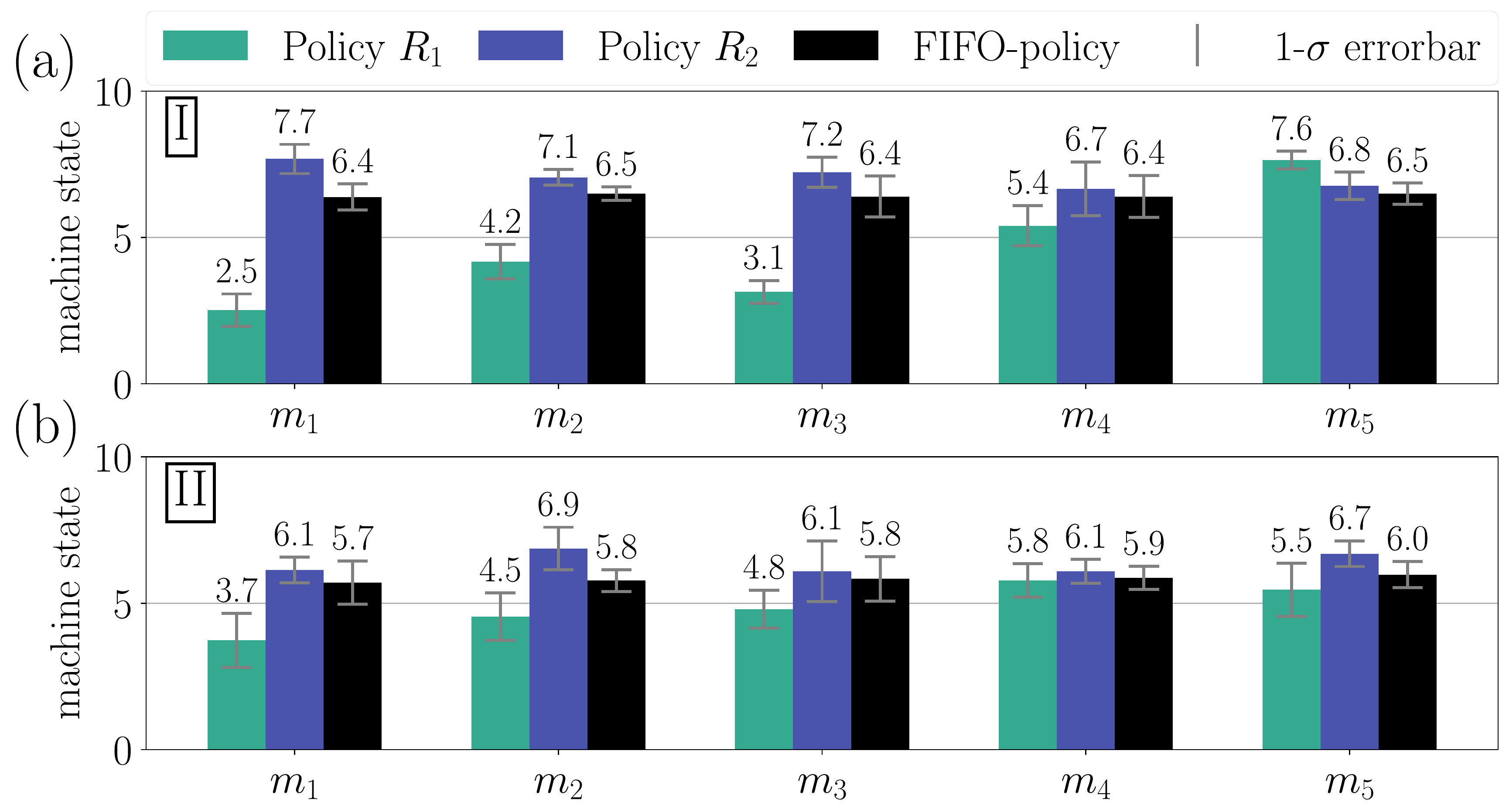}
     \caption{Average machine condition states of CBM actions in (a) configuration~\uproman{1} and (b) configuration~\uproman{2}.}
     \label{State_CBM}
\end{figure}

\begin{figure}[tb]
    \centering
    \includegraphics[width=\linewidth]{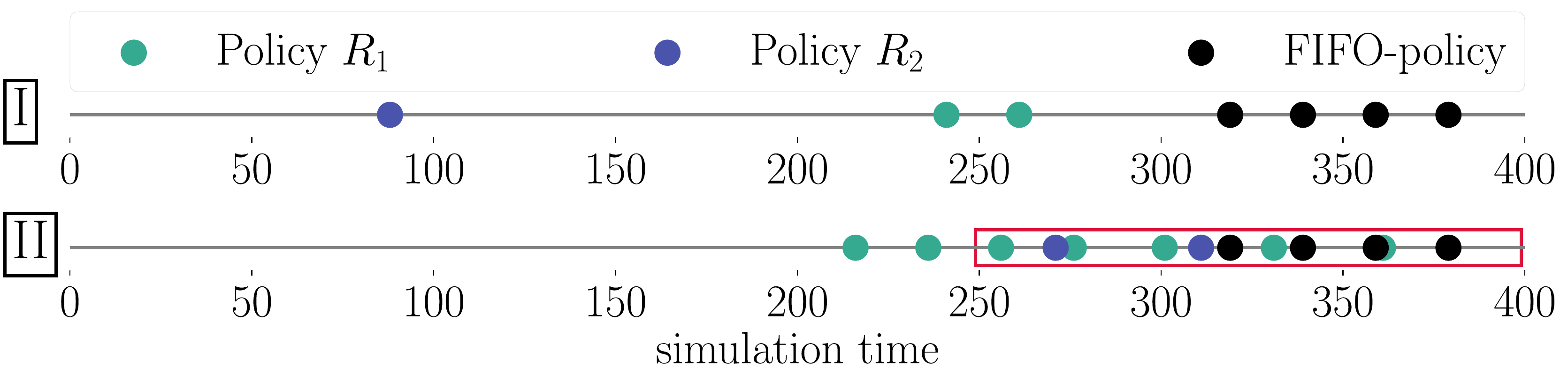}
     \caption{CM actions for a simulation episode in configuration~\uproman{1} and~\uproman{2}.}
     \label{timesteps_CM}
\end{figure}

The timing of the performed maintenance action is another dimension for evaluating the behavior of the learned maintenance scheduling policies. For preventive actions, the machine condition at which the CBM action is executed is crucial. In \Fig{State_CBM} the average machine condition at the time a CBM action is executed is displayed for configuration \uproman{1} and \uproman{2}. Policy $R_{2}$ schedules CBM actions at an average condition state of 6.1 to 7.7 whereas the average condition of the benchmark FIFO-policy is between 5.7 and 6.5. Therefore, policy $R_{2}$ waits longer until CBM actions are scheduled and makes the best use of the remaining useful life (RUL) of the single machines in both configurations. Policy $R_{1}$ in contrast schedules CBM actions at a condition state between 2.5 and 7.6 and does not use the RUL sufficiently. This phenomenon can be attributed to the fact that Policy $R_{1}$ never chooses the idle action and performs CBM actions instead. While the Policy $R_{2}$ performs on average 176 actions (46.1 CBM, 2.2 CM, 127.7 idle) in configuration~\uproman{1}, Policy $R_{1}$ only performs up to 73 actions (70.6 CBM, 2.4 CM) and never chooses any idle action, see Table~\ref{table:2}.

\begin{figure*}[tb]
    \centering
    \includegraphics[width=\textwidth]{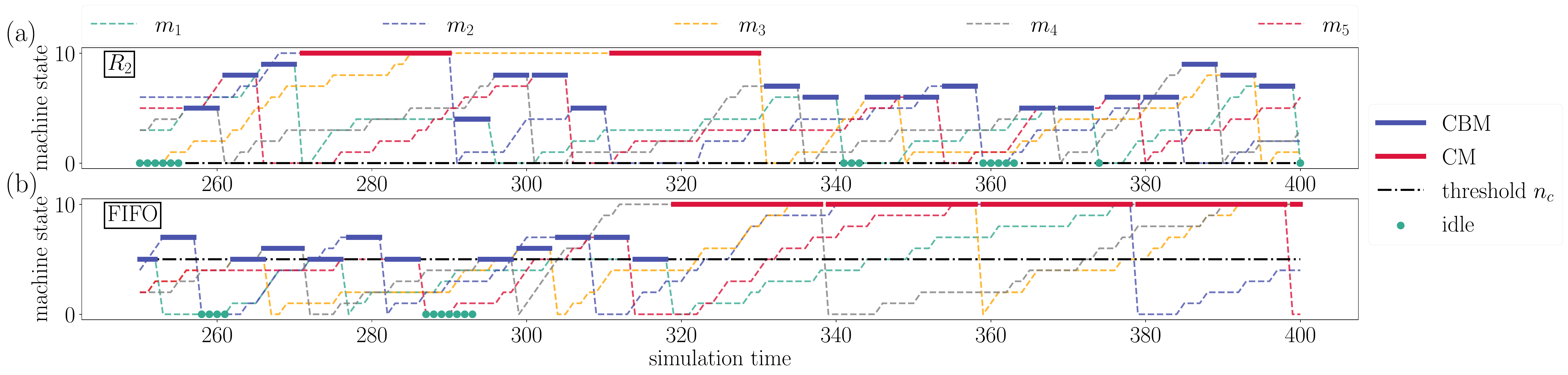}
    \caption{Detail view (red box in \Fig{timesteps_CM}) of the actions in the last 150 simulation time steps in configuration~\uproman{2}.}
    \label{detail_policy}
\end{figure*}

Another difference between the policies is the timing of performed corrective maintenance actions, which have to be executed when a machine breaks down. In \Fig{timesteps_CM} the performed CM actions for a representative episode are displayed over the simulation steps. Policy $R_{2}$ only uses one CM in configuration~\uproman{1} and two CM in configuration~\uproman{2} and therefore shows the lowest number of corrective actions. For the FIFO-policy, a maintenance holdup can be observed at the end of the simulation run. With the FIFO-policy, CM actions increasingly arise in both production configurations towards the end of the simulation. This behavior is not observed for policy $R_{2}$. In \Fig{detail_policy} the machine conditions and performed actions for the last 150 simulation time steps of configuration \uproman{2} in \Fig{timesteps_CM}, indicated by the red box, are shown. In order to avoid maintenance holdup situations, $R_{2}$ (\Fig{detail_policy}(a)) prioritizes necessary CBM actions. At $t=285$ $m_{3}$ breaks down. To prevent a series of CM actions, $R_{2}$ first maintains $m_{1}$, $m_{4}$, $m_{5}$, and $m_{2}$, which are also in higher condition states and only executes the CM action for $m_{3}$ after all other machines are in non-critical states. Since policy $R_{1}$ is not competitive it is not considered in the detailed analysis.


\section{Conclusion}

In this paper, reinforcement learning is used to learn condition-oriented maintenance scheduling strategies for a flow line system, which are evaluated in a synchronous and asynchronous configuration production system. The best-learned policy achieves better results than a benchmark FIFO-policy in both configurations. Modelling of the reward is crucial for both, getting the intended behavior and for understanding the learned policies. The evaluation of the policies shows that both RL-based policies do not show anomalous or odd behavior and the actions can be reconstructed and understood by human operators.
Further research will focus on methods to evaluate learned maintenance policies in a more general setting and on leveraging prior knowledge as well as model-based RL to apply maintenance scheduling for more complex production settings with multi-variant products towards the goal of using RL for maintenance scheduling in real-world applications.

\section*{Acknowledgment}

This work was supported by the Baden-Wuerttemberg Ministry for Economic Affairs, Labour and Housing (Project \frqq KI-Fortschrittszentrum LERNENDE SYSTEME\flqq).

The implementation of the simulation model and the performed experiments are available at: \url{https://github.com/ral94/rlcbm}

\balance

%

\end{document}